\documentclass[a4paper,fleqn]{cas-sc}

\usepackage[numbers,sort&compress]{natbib}

\begin{document}
\let\WriteBookmarks\relax

\shorttitle{CER for Image-Based Inspection}
\shortauthors{T.~Zhang et al.}

\title[mode=title]{Capacity-equilibrium reconstruction for multi-class anomaly detection in image-based inspection}

\author[1,3]{Tian Zhang}[orcid=0009-0006-4063-3664]
\ead{zhangtian2021@mail.nwpu.edu.cn}
\author[1]{Xian Feng}[orcid=0000-0002-5018-7171]\cormark[1]
\ead{fengxa@nwpu.edu.cn}
\author[2,3]{Tao Yang}[orcid=0000-0002-0714-2904]
\ead{yangtao107@nwpu.edu.cn}

\affiliation[1]{organization={School of Marine Science and Technology, Northwestern Polytechnical University},
city={Xi'an}, postcode={710072}, country={China}}
\affiliation[2]{organization={School of Automation, Northwestern Polytechnical University},
city={Xi'an}, postcode={710072}, country={China}}
\affiliation[3]{organization={Datago Information Technology},
city={Xi'an}, postcode={712000}, country={China}}

\cortext[1]{Corresponding author.}

\begin{abstract}
Multi-class unsupervised anomaly detection methods are trained only on normal samples and cover multiple products simultaneously with a single model. Such methods are particularly useful in industrial non-destructive testing where defect samples are scarce. Currently, multi-class anomaly detection based on feature reconstruction achieves the best performance. However, existing methods only evaluate the model performance based on the final detection accuracy, without analyzing the impact of decoder modifications on the reconstruction of normal and abnormal samples. In this paper, the effect of decoder capacity on detection performance is decomposed into two measurable quantities: reconstruction fidelity on normal images and anomaly sensitivity on abnormal images. When the within-population dispersion is fixed, the image-level AUROC monotonically increases with the sum of these two quantities. This is the capacity-equilibrium principle proposed in this paper. The Capacity-Equilibrium Reconstruction (CER) method uses Distant-Context Attention to improve sensitivity, Layer-wise Capacity Compensation to improve fidelity, and Sample-wise Hard-region Mining to balance the training of different samples. CER achieves the highest image-level AUROC on Uni-Medical, MVTec-AD, and VisA.
\end{abstract}


\maketitle

\section{Introduction}\label{sec:intro}

In the field of industrial non-destructive testing, inspectors are required to identify the samples with quality defects from a large number of input samples. However, during manual inspection, inspectors need to examine many similar images for a long time, which can lead to fatigue and result in missed detections or false alarms. As production lines and equipment expand, the number of images to be reviewed keeps increasing, and the types of samples to be inspected also become more diverse. Automated anomaly detection methods have received much attention. Supervised anomaly detection methods need many labeled defect samples for training. However, on the production line, most samples are defect-free products, and defect samples are relatively scarce. The defect shapes are diverse, making it difficult to ensure that all defect types are collected. This makes it difficult for supervised anomaly detection methods to generalize to unseen anomalies, resulting in missed detections. Unsupervised anomaly detection only uses normal samples for training and does not rely on any defect annotation. During inference, abnormal samples that deviate from the normal pattern are given high anomaly scores. Therefore, unsupervised anomaly detection methods are particularly suitable for industrial product quality inspection~\cite{bergmannMVTecAnomalyDetectionDataset2021,zouSPotDifferenceSelfSupervisedPreTraining2022}.

As the number of product categories increases, multiple types of samples need to be inspected simultaneously. In the multi-class anomaly detection task, most existing unsupervised methods train separately for each category. As the number of categories increases, not only does each category need to be trained separately, but the storage and runtime costs required during inference also increase proportionally. A more practical approach is to use a unified model to detect anomalies for all categories simultaneously, that is, multi-class anomaly detection~\cite{youUnifiedModelMulticlass2022}. Under this setting, the training set includes multiple categories of normal samples with significant appearance differences. On industrial production lines, there are circuit board samples with similar appearances but different circuit layouts. In these cases, single-class anomaly detection models regard the differences between the normal appearances of different categories as abnormal regions. For multi-class anomaly detection, the model needs to learn the normal patterns of each category without class labels. The boundary between normal and abnormal becomes more complex.

Currently, anomaly detection methods based on feature reconstruction achieve the best performance in multi-class anomaly detection. These methods utilize a frozen model pre-trained on large-scale general data to extract the features of the image. A trainable decoder is then used to learn and reconstruct the normal-pattern features of the frozen model. During the training of the decoder, only normal samples are used. Therefore, during inference, the features of normal images can be correctly reconstructed by the decoder, while the features of abnormal images cannot be reconstructed. The reconstruction error can be regarded as the anomaly score of the input sample~\cite{youUnifiedModelMulticlass2022,guoDinomalyLessMorePhilosophy2025}.

It can be seen that the core of the anomaly detection method based on feature reconstruction lies in the model structure and reconstruction capacity of the reconstruction decoder. The decoder needs to have a strong reconstruction capacity to effectively reconstruct normal samples of different categories. However, an excessively large decoder capacity will cause the decoder to learn an identical shortcut. That is, the abnormal region can also be reconstructed through self-attention neighborhood copying, resulting in the disappearance of its reconstruction error. When samples of multiple categories are trained simultaneously, due to the differences between categories, this over-generalization becomes more severe~\cite{youUnifiedModelMulticlass2022}. To address this, UniAD adds a fixed-window neighbor mask to the self-attention module to block the neighborhood copying of abnormal features and reduce the decoder's over-generalization to abnormal samples. Recently, Dinomaly has used linear attention and dropout to reduce the capacity of the decoder. However, these methods do not discuss the impact of the decoder's capacity on the reconstruction performance of abnormal and normal samples; they only use the final detection accuracy to verify the effectiveness of model improvement.

According to the principle of the feature reconstruction anomaly detection method, we find that increasing the decoder depth leads to better reconstruction of both normal and abnormal samples. The decoder structure optimization in this direction does not improve the performance of the detector. Therefore, this paper decomposes the decoder capacity into two measurable quantities: fidelity ($\mathrm{F}$) and sensitivity ($\mathrm{S}$). The fidelity is defined as the negative of the mean reconstruction error on normal images, which indicates whether the decoder can reconstruct the features of normal images well. The sensitivity is defined as the mean of the reconstruction error on abnormal images and measures the extent to which the decoder does not reconstruct abnormal regions. The sum of fidelity and sensitivity is the difference in the average reconstruction error between normal and abnormal images, which is referred to as the detection margin in this paper. When the dispersion of anomaly scores in normal and abnormal samples is fixed, the image-level AUROC rises monotonically with the detection margin. This is the capacity-equilibrium principle proposed in this paper. The effect of each module in the decoder on the detection performance can be divided into its impact on fidelity and sensitivity. The improvement of the decoder structure can only improve the detection performance if it increases the detection margin.

\begin{figure}[pos=htbp]
  \centering
  \includegraphics[width=\textwidth]{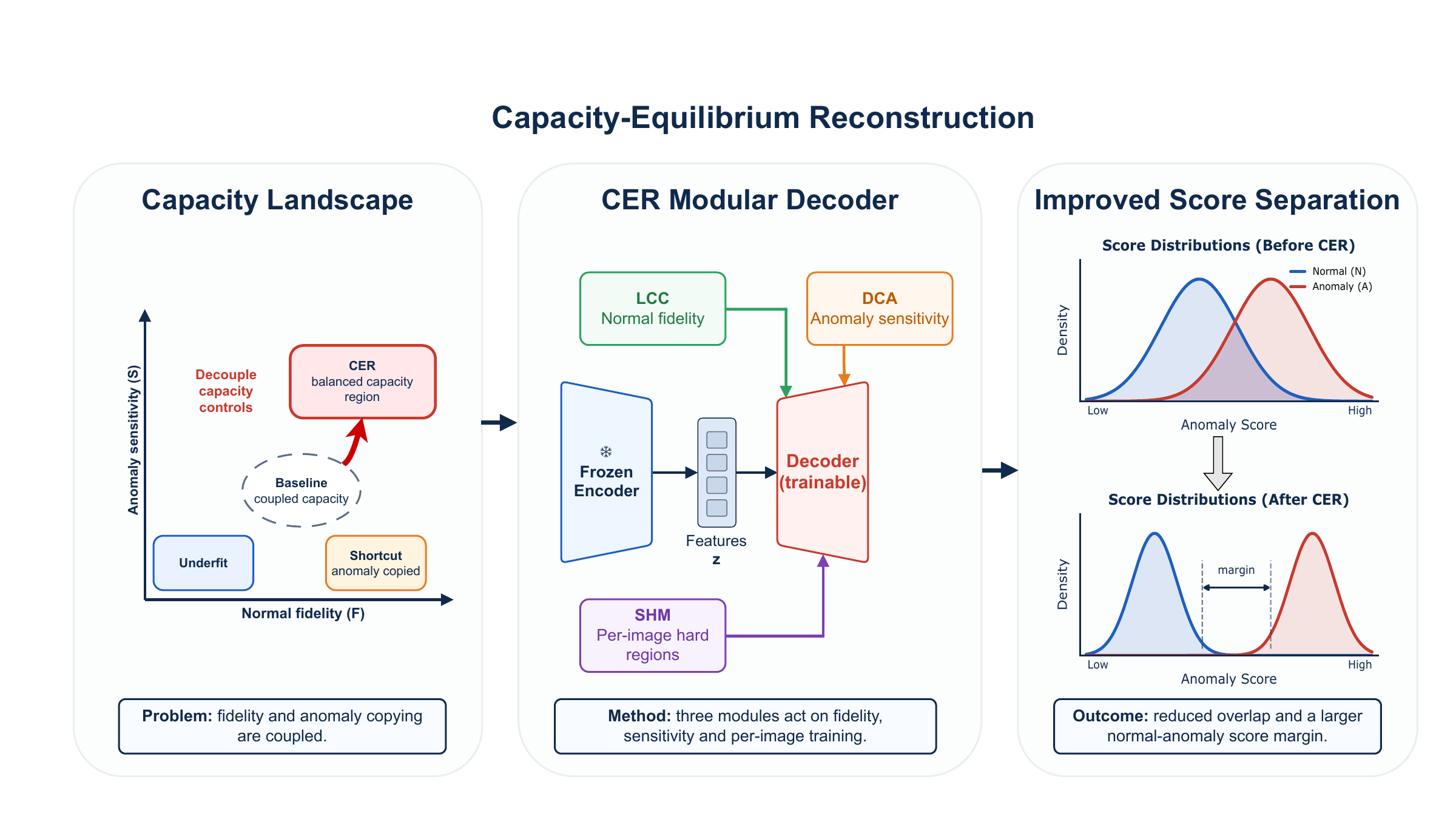}
  \caption{Illustration of the capacity-equilibrium principle and the CER method.
  Left: fidelity and sensitivity are coupled when only the capacity changes.
  Middle: the three modules of CER.
  Right: score distributions of normal and abnormal images before and after CER.}
  \label{fig:concept}
\end{figure}

Based on this principle, we propose the Capacity-Equilibrium Reconstruction (CER) method, which consists of three modules, as shown in Fig.~\ref{fig:concept}. First, we design the Distant-Context Attention (DCA) module to prevent abnormal region features from being copied. DCA uses a neighbor mask to block the correlation calculation near each token before the softmax of the self-attention module. This forces the self-attention module to reconstruct each token from more distant tokens. It can reduce the decoder's ability to reconstruct abnormal sample features, thereby improving the sensitivity. The size of the mask window gradually increases from shallow to deep within a narrow band to enhance the decoder's robustness to the window size. Second, the Layer-wise Capacity Compensation (LCC) module is proposed to enhance the decoder's ability to reconstruct normal samples. It adds a path that does not pass through self-attention in each block of the decoder. LCC first aggregates the input features of the entire image into a summary vector and broadcasts it to all tokens, and then performs a feedforward transformation separately for each position. This path supplements the capacity required for the decoder to reconstruct the normal features. Thus, LCC improves the fidelity of the decoder. Since the tokens at different positions do not exchange information, LCC avoids the shortcut for copying anomalies. Finally, to address uneven sample training under the multi-class setting, the Sample-wise Hard-region Mining (SHM) module is proposed. It sets a threshold for hard-region mining on each image, and each image uses the most difficult-to-reconstruct region for training. SHM does not require any class labels.

The main contributions of this paper are summarized as follows. First, the decoder capacity in the feature reconstruction anomaly detection model is decomposed into two measurable quantities: fidelity and sensitivity, which quantify the relationship between decoder capacity and detection performance. The capacity-equilibrium principle is proposed, and it is predicted that simply increasing the decoder capacity cannot effectively improve the detection accuracy of the model.

Second, we propose the CER method, which consists of three modules. DCA improves sensitivity, LCC improves fidelity, and SHM trains each image in the area where it is the most difficult to reconstruct without class labels. The contribution of each module to detection performance is analyzed using the capacity-equilibrium principle.

Third, we evaluate the CER method on multiple datasets. On the Uni-Medical dataset, the image-level AUROC reaches 87.61\%, exceeding the strongest baseline by 2.7 percentage points. On the MVTec-AD and VisA datasets, it also outperforms the baseline models, reaching 99.74\% and 98.86\% respectively.

This paper is organized as follows: Section~\ref{sec:related} introduces the related research work on anomaly detection; Section~\ref{sec:method} explains the capacity-equilibrium principle and elaborates on the proposed three modules; Section~\ref{sec:setup} presents the experimental settings and results on Uni-Medical, MVTec-AD, and VisA, and analyzes the effectiveness of each module; Section~\ref{sec:conclusion} concludes.

\section{Related work}\label{sec:related}

The reconstruction-based anomaly detection method utilizes an encoder and a decoder to learn the image patterns of normal samples. During testing, the reconstruction model cannot reconstruct the abnormal regions, so the reconstruction error can be used as the anomaly score of the input sample.

UniAD~\cite{youUnifiedModelMulticlass2022} proposed an anomaly detection model under the multi-class setting, using neighbor masked self-attention to suppress the identical shortcut problem. CNC~\cite{wangCNCCrossmodalNormality2025} argued that the abnormal regions were reconstructed well because the decoder over-generalized during multi-class training. They chose to utilize category-agnostic learnable text prompts and a gated mixture of experts to constrain the decoder features. Dinomaly~\cite{guoDinomalyLessMorePhilosophy2025} replaced the encoder with the frozen DINOv2 backbone and utilized the structural design to suppress the decoder's over-generalization problem. The state-space model decoder MambaAD~\cite{heMambaADExploringState2024}, the prototype query variant INP-Former~\cite{luoExploringIntrinsicNormal2025} and other methods~\cite{wangDeepFeatureClustering2025,luoFeatureShufflingRestoration2026} all optimize the decoder structure for multi-class settings. The neighbor mask of UniAD~\cite{youUnifiedModelMulticlass2022}, the memory module of MemAE~\cite{gongMemorizingNormalityDetect2019}, and the linear attention of Dinomaly~\cite{guoDinomalyLessMorePhilosophy2025} all adjust the reconstruction capacity of the decoder through specialized decoder structure design. However, these decoder structure designs only test the module effect based on the final detection accuracy and do not discuss whether the decoder can reconstruct normal samples without reconstructing abnormal samples. This paper converts this requirement into two measurable quantities, and uses them as the directions for improving the decoder to enhance the model's detection performance.

\section{Method}\label{sec:method}

The CER method we propose consists of one principle and three modules. Based on how anomaly scores are calculated in feature reconstruction methods, we decompose the capacity of the decoder into two measurable quantities: fidelity and sensitivity, and propose the capacity-equilibrium principle. According to this principle, we find that improving the decoder's ability to reconstruct the features of normal samples by increasing the depth alone will cause the fidelity and sensitivity to rise and fall respectively, canceling each other out. Therefore, the improvement of this type of decoder has a limited effect on anomaly detection performance. This paper designs three modules based on fidelity and sensitivity. In the experiments, we verify how the proposed modules affect fidelity and sensitivity, and ultimately improve the performance of the detector.

\begin{figure}[pos=htbp]
  \centering
  \includegraphics[width=\textwidth]{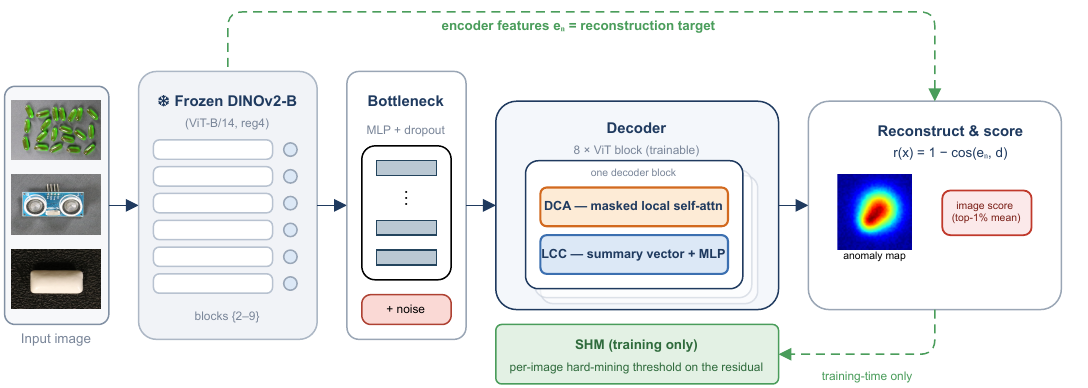}
  \caption{The overall framework of CER.
  DCA and LCC are placed in each decoder block, and SHM acts on the training loss.}
  \label{fig:arch}
\end{figure}

The overall structure of the proposed model is shown in Fig.~\ref{fig:arch}. Among the three proposed modules, DCA is placed within the self-attention of the decoder, LCC is attached to each decoder block, and SHM acts on the training loss. These three modules do not overlap with each other. The features of the input image are extracted using the frozen pre-trained model DINOv2-Base~\cite{oquabDINOv2LearningRobust2024}. We use the features of the 2nd to 9th blocks extracted by the pre-trained encoder. The 2nd to 5th blocks are combined as the shallow group, and the 6th to 9th blocks are combined as the deep group. These two fused features are the encoder features that the decoder needs to reconstruct. In addition, all the features from the 2nd to 9th blocks are linearly fused and averaged, and then passed through a bottleneck layer as the input of the decoder. The outputs of the eight decoder blocks are aggregated separately in two groups, the shallow and deep, as the reconstructed features of the decoder output. The difference between the reconstructed features of the decoder and the encoder features constitutes the anomaly score map. The mean of the top 1\% values in the anomaly score map is taken as the image-level anomaly score, and is denoted as the reconstruction residual $r(x)$.

\subsection{The capacity-equilibrium principle}\label{sec:method:ce}

This section decomposes the decoder capacity into two measurable quantities. It also shows that when the dispersion of the scores of normal and abnormal samples is fixed, the image-level AUROC depends on the sum of these two quantities. Let $f$ be the frozen encoder and $g$ be the decoder. $r(x) = \lVert f(x) - g(f(x)) \rVert$ is the image-level reconstruction residual of the input $x$, which is the mean value of the top 1\% positions of the score map. We denote the expected image-level anomaly scores on normal test samples and abnormal test samples as $\rho_n$ and $\rho_a$ respectively. The fidelity is defined as $\mathrm{F}(g) = -\rho_n$, and the sensitivity is defined as $\mathrm{S}(g) = \rho_a$, as shown in Eqs.~\eqref{eq:fidelity} and~\eqref{eq:sensitivity}.
\begin{align}
\mathrm{F}(g) \;&\triangleq\; -\rho_n \;=\; -\,\mathbb{E}_{x\sim p_n}\!\left[r(x)\right] \qquad (\text{fidelity}), \label{eq:fidelity}\\
\mathrm{S}(g) \;&\triangleq\; \phantom{-}\rho_a \;=\; \phantom{-\,}\mathbb{E}_{x\sim p_a}\!\left[r(x)\right] \qquad (\text{sensitivity}), \label{eq:sensitivity}
\end{align}
According to the definitions of fidelity and sensitivity, both of these values should be as large as possible. Here, $\mathrm{F}$ can be interpreted as the decoder's faithful reconstruction of normal samples, and $\mathrm{S}$ represents the decoder's failure to reconstruct the abnormal regions. The detection margin $\Delta = \rho_a - \rho_n = \mathrm{F}(g) + \mathrm{S}(g)$ is the sum of the two quantities. When the dispersion of the anomaly scores of normal and abnormal samples is fixed, the image-level AUROC can be expressed as a strictly increasing function of $\Delta$, as shown below.

First, assume the decision threshold is $t$, meaning that images with an anomaly score $r(x) > t$ will be classified as abnormal samples. Thus, the true positive rate can be written as $\mathrm{TPR}(t) = P\big(r(x_a) > t\big)$, and the false positive rate can be written as $\mathrm{FPR}(t) = P\big(r(x_n) > t\big)$. By varying the decision threshold $t$, the receiver operating characteristic (ROC) curve can be obtained. According to the probability form of AUROC~\cite{hanleyMeaningUseArea1982}, the area under the curve satisfies Eq.~\eqref{eq:auroc_prob}.
\begin{equation}\label{eq:auroc_prob}
\mathrm{AUROC} \;=\; \int_0^1 \mathrm{TPR}\, d(\mathrm{FPR})
\;=\; \int_{-\infty}^{\infty} P\big(r(x_a) > t\big)\, q_n(t)\, dt
\;=\; P\big(r(x_a) > r(x_n)\big),
\end{equation}
where $x_n \sim p_n$ and $x_a \sim p_a$ are independently sampled, and $q_n$ is the density of the anomaly score $r(x_n)$ for normal samples. The expression in the middle of Eq.~\eqref{eq:auroc_prob} calculates the conditional probability that the anomaly score exceeds the threshold $t$, and performs a weighted average over the probability distribution of the normal sample scores. That is, AUROC can be understood as the probability that the anomaly score of a randomly selected abnormal image is higher than that of a randomly selected normal image.

Second, the distribution of scores for normal samples and abnormal samples is rewritten in the form of a location-shift model, as shown in Eq.~\eqref{eq:location_shift}:
\begin{equation}\label{eq:location_shift}
r(x_n) = \rho_n + \sigma\,\varepsilon_1, \qquad r(x_a) = \rho_a + \sigma\,\varepsilon_2,
\end{equation}
where $\varepsilon_1$ and $\varepsilon_2$ are independent and identically distributed continuous noise variables with a mean of zero, and $\sigma$ is the common within-population scale parameter. We substitute Eq.~\eqref{eq:location_shift} into Eq.~\eqref{eq:auroc_prob}, and then divide both sides of the resulting inequality by $\sigma$. This gives us Eq.~\eqref{eq:auroc_margin}.
\begin{equation}\label{eq:auroc_margin}
\mathrm{AUROC} \;=\; P\big(\varepsilon_2 - \varepsilon_1 > -\Delta/\sigma\big) \;=\; H\big(\Delta/\sigma\big),
\end{equation}
where $H$ is the cumulative distribution function (CDF) of the difference $\varepsilon_2 - \varepsilon_1$. The second equality in Eq.~\eqref{eq:auroc_margin} is derived using the symmetry of the combined noise term. According to the definition of the CDF, the probability on the left side is equal to $1 - H(-\Delta/\sigma)$. Since $\varepsilon_1$ and $\varepsilon_2$ are independent and identically distributed, it follows that $\varepsilon_2 - \varepsilon_1$ and $\varepsilon_1 - \varepsilon_2$ have the same probability distribution. Due to this mirror symmetry, for any $u$, $H(-u) = 1 - H(u)$. Thus, $1 - H(-\Delta/\sigma)$ can be simplified to $H(\Delta/\sigma)$. AUROC directly depends on the standardized margin $\Delta/\sigma$. Since the CDF is strictly monotonically increasing, AUROC also increases monotonically with $\Delta/\sigma$.

Finally, consider the special case where the scores follow a Gaussian distribution. Given that the variance of the difference between two independent variables is equal to the sum of their individual variances, we have $r(x_a) - r(x_n) \sim \mathcal{N}(\Delta, 2\sigma^2)$. Eq.~\eqref{eq:auroc_margin} can thus be rewritten as Eq.~\eqref{eq:auroc_phi}, where $\Phi$ represents the CDF of the standard normal distribution.
\begin{equation}\label{eq:auroc_phi}
\mathrm{AUROC} \;=\; \Phi\!\left(\frac{\Delta}{\sqrt{2}\,\sigma}\right),
\end{equation}
Here, $\Delta/\sigma$ is the discriminability index $d'$ in the equal-variance binormal model~\cite{greenSignalDetectionTheory1966}. Thus, when the within-population dispersion is fixed, increasing $\Delta = \mathrm{F} + \mathrm{S}$ can improve the AUROC of the detection model. This paper takes $\Delta$ as the design target. The improvement of each module can be verified on the empirical $(\mathrm{F},\mathrm{S})$ plane. The training of the anomaly detection model only uses normal samples, and $\mathrm{S}$ is the expectation on the abnormal samples. $\Delta$ is only used to evaluate the modules after training. No abnormal samples are introduced during training.

$\mathrm{F}$ and $\mathrm{S}$ are coupled. Increasing the decoder depth alone will cause $\mathrm{F}$ to increase while $\mathrm{S}$ decreases. Simply increasing the number of decoder blocks cannot effectively improve the performance of the anomaly detection model. This is because when the decoder uses local copying to improve the reconstruction of normal patches, the abnormal patches will also be faithfully reconstructed using the same path. Reducing the capacity of the decoder can effectively prevent the decoder from reconstructing abnormal features, but it also reduces the fitting of normal samples, resulting in a gradual collapse of the discrimination between the anomaly scores of the two types of samples. The CER method adds enhancement modules in both the fidelity and sensitivity directions. DCA prohibits each token in the self-attention layer from obtaining the information of its adjacent tokens, reducing the reconstruction of abnormal regions, and improves the sensitivity. LCC follows a path that does not involve self-attention, and enhances the reconstruction quality of normal samples, thereby improving the fidelity. Finally, SHM sets a threshold for hard-region mining on each sample, allowing each image to continue training in the area where it is most difficult to reconstruct, improving the reconstruction difference between normal and abnormal samples.

\subsection{Distant-Context Attention (DCA)}\label{sec:method:dca}

In the self-attention module, tokens tend to obtain information from their own neighborhood features for reconstruction. This local copying phenomenon allows the decoder to learn a mapping path during training. This path is also known as the identical shortcut. To address this issue, we design the DCA module, which prohibits each patch token from attending to its own neighborhood, thereby improving the decoder's sensitivity.

The two-dimensional patch tokens are numbered as $\operatorname{patch}(i)$, and $\mathbf{q}_i$ and $\mathbf{k}_j$ represent the query vector of token $i$ in an attention head and the key vector of token $j$, respectively. Before performing row-wise softmax normalization, spatial masking is applied to the attention matrix, as shown in Eq.~\eqref{eq:dca}.
\begin{equation}\label{eq:dca}
A_{ij} \;=\;
\begin{cases}
-\infty, & \lVert \operatorname{patch}(i) - \operatorname{patch}(j) \rVert_\infty \le (N-1)/2, \\[2pt]
\mathbf{q}_i^\top \mathbf{k}_j / \sqrt{d_h}, & \text{otherwise,}
\end{cases}
\end{equation}
When the Chebyshev distance between tokens $i$ and $j$ is within $(N-1)/2$, their attention values are set to $-\infty$. Attention values at other positions are still calculated according to $\mathbf{q}_i^\top \mathbf{k}_j / \sqrt{d_h}$, and the subsequent softmax is calculated row-wise on the remaining positions. Here, $d_h$ is the feature dimension of a single self-attention head. When $N$ is an odd number, the region covered by $\lVert \operatorname{patch}(i) - \operatorname{patch}(j) \rVert_\infty \le (N-1)/2$ is the $N\times N$ square centered on the query token $i$. The neighborhood features are masked and excluded from the softmax normalization, so each token can only obtain information from more distant context features for feature reconstruction. The copying path of tokens is removed, and the decoder's sensitivity is thereby improved.

\paragraph{Multi-scale form.}
This paper extends the single mask size value to be set separately by block to enhance the robustness of window size parameter selection. We denote $A(N)$ as the masked attention matrix of the window size $N$ in Eq.~\eqref{eq:dca}. The decoder's $\ell$th block uses $A(N_\ell)$. From the shallow blocks to the deep blocks of the decoder, $N_\ell$ takes values from small to large within a narrow band $B$, as shown in Eq.~\eqref{eq:dca_ms}.
\begin{equation}\label{eq:dca_ms}
A^{(\ell)} \;=\; A\!\left(N_{\ell}\right),
\qquad N_{\ell}\in B,\quad N_{1}\le N_{2}\le\dots\le N_{L}.
\end{equation}
By using multiple window sizes, not only can strongly correlated tokens be blocked from the reconstruction, but also the performance of the model does not depend on a specific window size. The same narrow-band window size configuration can be directly used for all three benchmarks, without the need to tune the global window size individually on different datasets.

\subsection{Layer-wise Capacity Compensation (LCC)}\label{sec:method:lcc}

Under the multi-class setting, a decoder needs to simultaneously reconstruct the normal patterns of all categories. The neighbor mask in Section~\ref{sec:method:dca} further limits the reconstruction performance of the decoder. Therefore, the decoder needs to increase its capacity to improve its fidelity. According to the analysis in Section~\ref{sec:method:ce}, the increased capacity through the self-attention path also enhances local copying while improving the reconstruction of normal samples. To address this issue, we add an LCC module before the masked self-attention module from Section~\ref{sec:method:dca}. The LCC module calculates the summary vector of the input image and broadcasts it to all feature tokens. Each token is subsequently reconstructed based on this summary vector. During reconstruction, the token features are independent of each other, bypassing the information exchange of self-attention. The capacity of the decoder is enhanced without introducing the identical shortcut. Each block in the decoder has the same structure.

The summary vector $\mathbf{b}$ is calculated only once for each input sample, and its result is independent of the position of the token. Using the LCC module, each token in the decoder block can reconstruct the image features with the summary vector $\mathbf{b}$ of the block. This summary vector is sample-dependent and adaptive. When the normal data manifolds vary greatly, this adaptation can enhance the reconstruction quality of the decoder and reduce the scale of the residuals. That is, the decoder can identify which normal data manifold the current input image belongs to. The tokens, using the summary vector, can better utilize the distant context for feature reconstruction. Combined with the mask structure, the features of the abnormal area cannot copy themselves through these means. This not only increases the capacity of the decoder but also avoids the problem of the identical shortcut that is prone to occur in the self-attention module.

DCA and LCC together form an improved decoder block, and the decoder is constructed by stacking this block. DCA limits the reconstruction capacity to enhance sensitivity, while LCC supplements the reconstruction capacity to improve fidelity.

\subsection{Sample-wise Hard-region Mining (SHM)}\label{sec:method:shm}

Multi-class anomaly detection models detect anomalies on samples of different categories and with diverse appearances. The model needs to remain sensitive to each image. Hard example mining was first proposed in~\cite{shrivastavaTrainingRegionBased2016}, and later ReContrast and Dinomaly, among others~\cite{guoReContrastDomainSpecific2023,guoDinomalyLessMorePhilosophy2025}, applied it to the reconstruction loss. The hard-mining global cosine loss focuses on the most difficult-to-reconstruct features in the training set. The aim is to enable the decoder to discover global feature patterns during training, avoiding excessive focus on reconstruction details. The hard-mining global cosine loss sets an error threshold in a mini-batch, and features with reconstruction errors exceeding the threshold are normally backpropagated with gradients, while the gradients for features below the threshold are multiplied by a factor $\gamma$. Only the most difficult points to reconstruct update the decoder parameters with full gradients. The easier-to-reconstruct feature points only participate in the update with a very small weight. The hard-mining global cosine loss selects an error threshold for the entire mini-batch and applies it to each sample. Which areas in the sample can update the decoder parameters depends on how many difficult-to-reconstruct features are present in this sample. However, in multi-class training, the error scales of each category vary greatly, and there will always be a portion of samples that are easier to reconstruct. Therefore, if most regions of a sample have low reconstruction errors, only a small portion of the regions in that sample can allow the decoder to update normally. As the model training progresses, this bias accumulates and is directly reflected in the anomaly scores at inference. The image-level anomaly score of the input sample is defined as the mean of the reconstruction residuals at the top 1\% positions, which is the region of the image that is the most difficult to reconstruct. For images with defects but easy to reconstruct, the residuals of the defects must be higher than those of the other regions of the image to be reflected in the score. However, according to the definition of the hard-mining global cosine loss, the regions used for scoring in easy-to-reconstruct samples are actually lacking in training.

To address this issue, the SHM module in CER is used to calculate the loss function. Among the $M$ features extracted from a single image, let $d_i$ represent the cosine distance between the encoder feature and its reconstruction at the $i$-th point, where $i = 1, \dots, M$. The error threshold $\theta$ is taken as the $p$-th quantile of the cosine distances within this image, $\theta = \mathrm{Quantile}_p(\{d_i\})$. During the gradient update, the gradient scaling factor for the $i$-th point is as shown in Eq.~\eqref{eq:shm}.
\begin{equation}\label{eq:shm}
g_i =
\begin{cases}
1, & d_i \ge \theta,\\
\gamma, & d_i < \theta,
\end{cases}
\end{equation}
This approach enables the update of decoder parameters in the most challenging regions of the sample, even for samples with relatively easy-to-reconstruct features overall. Let $M_\theta$ represent the number of points in an image where $d_i \ge \theta$. The proportion of points with full gradients in each sample is fixed, as shown in Eq.~\eqref{eq:shm_inv}.
\begin{equation}\label{eq:shm_inv}
\frac{M_{\theta}}{M} \;=\; 1-p
\qquad\text{for every image}.
\end{equation}
During training, the areas that SHM focuses on learning in each sample are precisely the regions that are most crucial for discrimination during inference. The SHM module avoids the problem that some easy-to-reconstruct samples or categories in the training process are not adequately trained, while also retaining the decoder's ability to capture the global normal feature patterns.

\section{Experiments}\label{sec:setup}

This section first introduces the experimental setup. Then, the proposed method is compared with existing methods on the publicly available datasets Uni-Medical, MVTec-AD, and VisA. Next, an ablation experiment is conducted on the dataset with the highest performance improvement (Uni-Medical) and the influence of each module on the decoder capacity is analyzed. Finally, the parameter count and inference time are presented.

\subsection{Experimental setup}\label{sec:setup:exp}

\subsubsection{Datasets}\label{sec:setup:data}

This paper evaluates the CER method on three public benchmarks, namely Uni-Medical, MVTec-AD, and VisA. The numbers of classes and samples in each dataset are summarized in Table~\ref{tab:datasets}.

\begin{table}[tbp]
\centering
\caption{The numbers of categories and samples of the datasets used in the experiments.}
\label{tab:datasets}
\begin{tabular}{lcccc}
\toprule
Dataset & Categories & Train normal & Test normal & Test defect \\
\midrule
MVTec-AD~\cite{bergmannMVTecAnomalyDetectionDataset2021} & 15 & $3629$ & $467$ & $1258$ \\
VisA~\cite{zouSPotDifferenceSelfSupervisedPreTraining2022} & 12 & $8659$ & $962$ & $1200$ \\
Uni-Medical~\cite{baoBMADBenchmarksMedical2024,zhangComprehensiveLibraryBenchmarking2024} & 3 & $13339$ & $2514$ & $4499$ \\
\bottomrule
\end{tabular}
\end{table}

MVTec-AD~\cite{bergmannMVTecAnomalyDetectionDataset2021} and VisA~\cite{zouSPotDifferenceSelfSupervisedPreTraining2022} are commonly used datasets in industrial anomaly detection, with 15 and 12 classes respectively. In each dataset, the model is evaluated under the unified multi-class setting, using a single model to cover all categories in the dataset.

Uni-Medical~\cite{baoBMADBenchmarksMedical2024,zhangComprehensiveLibraryBenchmarking2024} is an open multi-class medical anomaly detection dataset, containing 13339 training images and 7013 test images. The samples include three categories: brain MRI~\cite{baidRSNAASNRMICCAIBraTS2021}, liver CT~\cite{landmanMICCAIMultiAtlasLabeling2015,bilicLiverTumorSegmentation2023}, and retinal OCT~\cite{huAutomatedSegmentationMacular2019}. It provides a low-contrast grayscale image testbed for the reconstruction capacity problem proposed in this paper.

\subsubsection{Metrics}\label{sec:setup:metrics}

This paper reports the average values of image-level and pixel-level AUROC for different methods, which are denoted as I-AUROC and P-AUROC in the tables. AUROC is the area under the receiver operating characteristic curve. In Section~\ref{sec:method:ce}, its equivalent probability form is also provided, which serves as the basis for the theoretical analysis in this paper. Following the baseline method Dinomaly~\cite{guoDinomalyLessMorePhilosophy2025}, we additionally report AP and F1-max on the Uni-Medical dataset. AP is the area under the precision-recall curve, and F1-max is the maximum F1 value among all decision thresholds. All results are averaged over all categories of the dataset.

\subsubsection{Implementation details}\label{sec:setup:impl}

The encoder uses the frozen pre-trained model DINOv2-Base with register tokens~\cite{darcetVisionTransformersNeed2024}. The decoder is composed of 8 stacked decoder blocks containing DCA and LCC, with an embedding dimension of 768. The model is trained on an NVIDIA RTX~4090. In Tables~\ref{tab:industrial} and~\ref{tab:medical}, the performance of CER and other methods is compared, where the CER results are the mean and standard deviation of three random seeds. The ablation experiments use a fixed seed to compare the role of each module.

\subsection{Comparison with existing methods}\label{sec:results:compare}

This paper compares the CER method with other multi-class anomaly detection methods on public datasets. The methods compared include RD4AD~\cite{dengAnomalyDetectionReverse2022}, UniAD~\cite{youUnifiedModelMulticlass2022}, DiAD~\cite{heDiffusionbasedFrameworkMulticlass2024}, ViTAD~\cite{zhangExploringPlainViTFeatures2024}, MambaAD~\cite{heMambaADExploringState2024}, and Dinomaly~\cite{guoDinomalyLessMorePhilosophy2025} based on reconstruction, as well as the SimpleNet~\cite{liuSimplenetSimpleNetwork2023} method based on embedding.

\begin{table}[tbp]
\centering
\caption{Comparison with existing methods on MVTec-AD and VisA under the multi-class setting (\%).
  CER: mean$\pm$std over three seeds.
  Best results in bold.
  The results of other methods are cited from~\cite{guoDinomalyLessMorePhilosophy2025}.}
\label{tab:industrial}
\begin{tabular}{lcccc}
\toprule
 & \multicolumn{2}{c}{MVTec-AD} & \multicolumn{2}{c}{VisA} \\
\cmidrule(lr){2-3}\cmidrule(lr){4-5}
Method & I-AUROC & P-AUROC & I-AUROC & P-AUROC \\
\midrule
RD4AD~\cite{dengAnomalyDetectionReverse2022}           & 94.6 & 96.1 & 92.4 & 98.1 \\
SimpleNet~\cite{liuSimplenetSimpleNetwork2023}         & 95.3 & 96.9 & 87.2 & 96.8 \\
UniAD~\cite{youUnifiedModelMulticlass2022}             & 96.5 & 96.8 & 88.8 & 98.3 \\
DiAD~\cite{heDiffusionbasedFrameworkMulticlass2024}    & 97.2 & 96.8 & 86.8 & 96.0 \\
ViTAD~\cite{zhangExploringPlainViTFeatures2024} & 98.3 & 97.7 & 90.5 & 98.2 \\
MambaAD~\cite{heMambaADExploringState2024}             & 98.6 & 97.7 & 94.3 & 98.5 \\
Dinomaly~\cite{guoDinomalyLessMorePhilosophy2025}      & 99.6 & 98.4 & 98.7 & 98.7 \\
CER (ours) & $\mathbf{99.74{\pm}0.01}$ & $\mathbf{98.43{\pm}0.01}$ & $\mathbf{98.86{\pm}0.06}$ & $\mathbf{98.97{\pm}0.01}$ \\
\bottomrule
\end{tabular}
\end{table}

\begin{table}[tbp]
\centering
\caption{Comparison with existing methods on Uni-Medical under the multi-class setting (\%).
  CER: mean$\pm$std over three seeds.
  Best results in bold.
  The results of other methods are cited from~\cite{guoDinomalyLessMorePhilosophy2025} or ADer~\cite{zhangComprehensiveLibraryBenchmarking2024}.}
\label{tab:medical}
\begin{tabular}{lcccccc}
\toprule
 & \multicolumn{3}{c}{Image-level} & \multicolumn{3}{c}{Pixel-level} \\
\cmidrule(lr){2-4}\cmidrule(lr){5-7}
Method & AUROC & AP & F1-max & AUROC & AP & F1-max \\
\midrule
RD4AD~\cite{dengAnomalyDetectionReverse2022}              & 76.1 & 75.3 & 78.2 & 96.5 & 38.3 & 39.8 \\
SimpleNet~\cite{liuSimplenetSimpleNetwork2023}            & 77.5 & 77.7 & 76.7 & 94.3 & 34.4 & 36.0 \\
UniAD~\cite{youUnifiedModelMulticlass2022} & 79.2 & 75.9 & 77.1 & 96.6 & 38.7 & 40.9 \\
DiAD~\cite{heDiffusionbasedFrameworkMulticlass2024}        & 78.8 & 77.2 & 77.7 & 95.8 & 34.2 & 35.5 \\
ViTAD~\cite{zhangExploringPlainViTFeatures2024}    & 81.8 & 80.7 & 80.0 & 97.1 & 48.3 & 48.2 \\
MambaAD~\cite{heMambaADExploringState2024} & 83.9 & 80.8 & 81.9 & 96.8 & 45.8 & 47.5 \\
Dinomaly~\cite{guoDinomalyLessMorePhilosophy2025}          & 84.9 & 84.1 & 81.0 & 96.8 & 51.7 & 52.1 \\
\midrule
CER (ours)                         & $\mathbf{87.61\pm0.09}$ & $\mathbf{87.16\pm0.11}$ & $\mathbf{82.21\pm0.25}$ & $\mathbf{97.23\pm0.10}$ & $\mathbf{56.58\pm1.03}$ & $\mathbf{54.96\pm0.81}$ \\
\bottomrule
\end{tabular}
\end{table}

The image-level and pixel-level AUROC of the CER method on the standard industrial anomaly detection datasets MVTec-AD and VisA are shown in Table~\ref{tab:industrial}. The image-level AUROC of existing methods on these two datasets has already approached the upper limit. The image-level AUROC of CER on MVTec-AD and VisA reaches 99.74\% and 98.86\% respectively, exceeding the 99.6\% and 98.7\% of the baseline method Dinomaly. The pixel-level AUROC of CER is 98.43\% and 98.97\%, also surpassing other methods.

To test the generalization ability of CER on cross-domain grayscale data, we evaluate the detection performance of CER and other methods on the Uni-Medical dataset. This dataset is significantly different from the natural image domain used by the pre-trained model, and all samples are low-contrast images. The results are shown in Table~\ref{tab:medical}. The improvement of CER on the Uni-Medical dataset is more pronounced, with image-level AUROC, AP, and F1-max reaching $87.61\pm0.09$\%, $87.16\pm0.11$\%, and $82.21\pm0.25$\% respectively, exceeding all other methods. The image-level AUROC of CER is 2.7 points higher than that of the Dinomaly method, which uses the same pre-trained encoder and input size.

In summary, the image-level AUROCs of CER on the three public benchmarks all exceed those of other methods. Even when using an encoder pre-trained on natural images, CER also achieves the best performance on medical grayscale images with significant distribution shift from natural images.

\subsection{Ablation and capacity analysis}\label{sec:results:ce_evidence}

\subsubsection{Module-wise ablation}\label{sec:results:loo}

We remove the Distant-Context Attention (DCA), Layer-wise Capacity Compensation (LCC), and Sample-wise Hard-region Mining (SHM) modules one by one from the complete configuration to verify the effectiveness of each module. The results are shown in Table~\ref{tab:ablation_loo}. After removing the DCA module, the image-level AUROC decreases from 87.69 to 84.08, a reduction of 3.61. This is because after removing the neighbor mask, the abnormal regions are reconstructed through the decoder's local copying, which reduces the decoder's sensitivity. After removing the LCC module, the decoder's capacity decreases, and the image-level AUROC drops to 86.53. This is because the appearance differences among samples of different categories in the dataset are extremely large. The decoder does not have sufficient capacity to reconstruct the normal image features of each category, resulting in a decline in the fidelity of the decoder. The dispersion $\sigma$ of the anomaly scores also increases from 1.92 to 2.01. The SHM module mines the hard regions of the sample in the training phase, and the image-level AUROC is reduced by 0.26 points after removing this module. After removing all the modules, the image-level AUROC decreases by 5.6 points to 82.09, and at this point, the decoder is a model with an identical shortcut. Here, the decoder uses the standard self-attention module, not the linear self-attention module used in Dinomaly~\cite{guoDinomalyLessMorePhilosophy2025}. The experimental results are consistent with our theoretical prediction.

\begin{table}[t]
\centering
\caption{Ablation study on Uni-Medical.
  $\mathrm{F}$ and $\mathrm{S}$ follow Eqs.~\eqref{eq:fidelity} and~\eqref{eq:sensitivity} with the class-balanced means of $r(x)$ over the normal and the abnormal test images, $\Delta=\mathrm{F}+\mathrm{S}$ is the detection margin, and $\sigma$ is the within-population scale parameter, the square root of the class-balanced mean of the normal and abnormal score variances.
  All four are in units of $10^{-2}$.}
\label{tab:ablation_loo}
\small
\begin{tabular}{l c c c c c c c}
\toprule
Configuration & $\mathrm{F}$ & $\mathrm{S}$ & $\Delta$ & $\sigma$ & $\Delta/\sigma$ & I-AUROC & Change \\
\midrule
CER (full)               & $-8.76$ & $11.91$ & $\mathbf{3.15}$ & $1.92$ & $\mathbf{1.64}$ & $\mathbf{87.69}$ & --- \\
\quad $-\,$LCC                      & $-9.41$ & $12.50$ & $3.10$ & $2.01$ & $1.54$ & $86.53$ & $-1.16$ \\
\quad $-\,$DCA (no mask)            & $-7.42$ & $9.64$  & $2.22$ & $1.62$ & $1.36$ & $84.08$ & $-3.61$ \\
\quad $-\,$SHM                      & $-8.59$ & $11.71$ & $3.12$ & $1.92$ & $1.62$ & $87.43$ & $-0.26$ \\
\quad all three removed             & $-8.88$ & $10.82$ & $1.94$ & $1.54$ & $1.26$ & $82.09$ & $-5.60$ \\
\bottomrule
\end{tabular}
\end{table}

\subsubsection{Fidelity--sensitivity plane}\label{sec:results:plane}

\begin{figure}[pos=htbp]
  \centering
  \includegraphics[width=0.8\columnwidth]{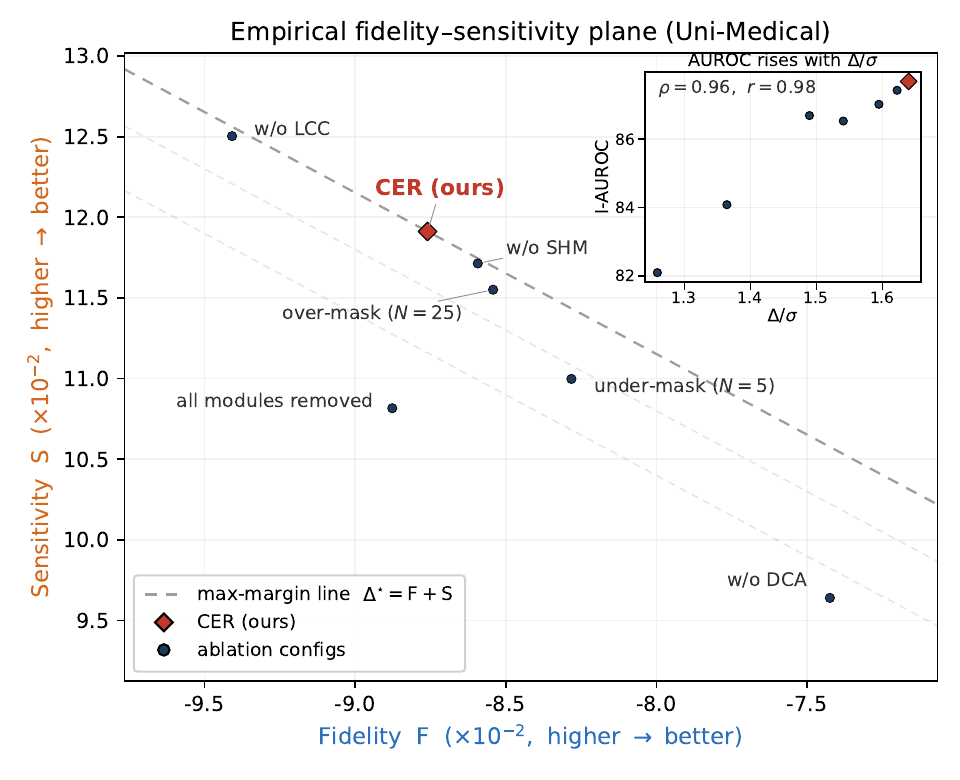}
  \caption{Model configurations on the $(\mathrm{F},\mathrm{S})$ plane of Uni-Medical.
  Points: the configurations in Table~\ref{tab:ablation_loo} and the two single-window configurations ($N = 5$ and $N = 25$).
  Gray dashed lines: equal $\Delta=\mathrm{F}+\mathrm{S}$.
  Inset: image-level AUROC against $\Delta/\sigma$.}
  \label{fig:er_plane}
\end{figure}

In Section~\ref{sec:method:ce}, we convert the performance evaluation of the detector into the detection margin, and then break down the detection margin into fidelity and sensitivity. To verify this decomposition, we plot the results obtained under different model configurations on the $(\mathrm{F},\mathrm{S})$ plane according to Eqs.~\eqref{eq:fidelity} and~\eqref{eq:sensitivity}. The model configurations include the five ablation test configurations mentioned in Table~\ref{tab:ablation_loo}, as well as the two single-window test configurations of $N = 5$ and $N = 25$. The results are shown in Fig.~\ref{fig:er_plane}. Among the seven configurations, the complete CER configuration achieves the largest detection margin $\Delta = \mathrm{F} + \mathrm{S}$. The other configurations also verify the capacity-equilibrium principle proposed in this paper. Removing LCC significantly improves the sensitivity of the decoder, but the model's fit to normal samples also drops. This increases the dispersion of the anomaly scores, and $\Delta/\sigma$ decreases accordingly. This leads to a decline in the model's performance. Removing DCA improves the decoder's ability to reconstruct normal features, but it also causes a serious identical shortcut problem. This results in a significant decrease in sensitivity, leading to a loss of model performance. As the mask window size gradually increases, the sensitivity increases and the fidelity decreases. When the window size increases to 25, the tokens get no useful context information. The reconstruction process relies only on residual connections and the MLP, so that both abnormal patches and normal patches are reconstructed better. The model's sensitivity decreases, the fidelity slightly recovers, and the detection margin becomes smaller. SHM trains each image in its most difficult area, and its $\Delta/\sigma$ increases from 1.62 to 1.64.

In addition, the relationship between the model's image-level AUROC and $\Delta/\sigma$ is shown in the inset of Fig.~\ref{fig:er_plane}. The results are consistent with the prediction of Eq.~\eqref{eq:auroc_phi}. The image-level AUROC increases with $\Delta/\sigma$. The Spearman correlation coefficient is 0.96 and the Pearson correlation coefficient is 0.98. In summary, we verify the capacity-equilibrium principle proposed in Section~\ref{sec:method:ce} through experiments, and also analyze the mechanism by which the proposed modules improve the model performance.

\subsection{Parameter count and inference time}\label{sec:results:efficiency}

This section evaluates the parameter count and inference time of CER. The results are shown in Table~\ref{tab:efficiency}. In CER, only the bottleneck layer and the decoder are involved in training, and the trainable parameters are 122.9M. During inference, including the DINOv2-Base feature extraction backbone, the deployed model has a total of 209.5M parameters. The model is a multi-class anomaly detection model, and the trainable parameters are shared among all classes. The inference parameter count does not increase with the number of covered classes. We test the inference speed of CER on an RTX~4090 with a batch size of 16 and an input size of 392. Processing one image takes an average of 18.3~ms, and the model processes 54.7 images per second. Therefore, in industrial inspection, the inference time will not become the bottleneck of first-round screening.

\begin{table}[tbp]
\centering
\caption{Computational cost of CER.
  The inference speed is tested on an RTX~4090 with a batch size of 16 and an input size of $392\times392$.}
\label{tab:efficiency}
\begin{tabular}{cccc}
\toprule
Trainable params (M) & Total params (M) & Time per image (ms) & Images per second \\
\midrule
$122.9$ & $209.5$ & $18.3$ & $54.7$ \\
\bottomrule
\end{tabular}
\end{table}

\section{Conclusion}\label{sec:conclusion}

This paper proposes a multi-class anomaly detection method, CER, based on the capacity-equilibrium principle. We introduce the capacity-equilibrium principle into the anomaly detection model based on feature reconstruction, splitting the capacity of the decoder into fidelity and sensitivity, and defining the sum of the two as the detection margin. The principle states that when the within-population dispersion is fixed, the image-level AUROC monotonically increases with the detection margin. The three modules of CER (DCA, LCC and SHM) improve the ratio $\Delta/\sigma$ between the detection margin and the dispersion of the anomaly scores from different perspectives, thereby enhancing the detection performance of the model. The DCA module adds multi-scale neighbor masks in the self-attention module of the decoder to cut off the copying path in feature reconstruction, thereby improving the sensitivity. LCC introduces a summary vector before the self-attention module to allow each token to learn the normal data manifold and reconstruct the features. This further provides the required capacity for the decoder to reconstruct the normal samples, enhances the fidelity, and narrows the dispersion of the anomaly scores. SHM sets the threshold for hard-region mining per image, allowing each image to be trained in the region where it is the most difficult to reconstruct. The performance of the model is tested on three datasets with different scenarios, and the results show that CER outperforms all other methods in image-level AUROC.

Although CER performs well in terms of image-level AUROC, it still has some limitations. Calculating sensitivity requires labeled abnormal samples, so the detection margin can only be used for analyzing the role of each module after training and cannot be used as a training target. The capacity-equilibrium principle is derived under the assumption that the score distributions of normal and abnormal samples only have location shifts, and $\Delta/\sigma$ is an approximate prediction of AUROC. In future work, we plan to extend the decomposition of fidelity and sensitivity to other pre-trained encoders and the larger-scale Real-IAD dataset~\cite{wangRealIADRealWorldMultiView2024}.

\bibliographystyle{elsarticle-num}
\bibliography{references}

\end{document}